\documentclass[letterpaper, 10 pt, conference]{ieeeconf}  

\IEEEoverridecommandlockouts                              

\let\labelindent\relax
\usepackage{enumitem}
\usepackage{times}
\usepackage{environ} 
\usepackage{graphicx}
\usepackage{amsmath}
\usepackage{amssymb}
\usepackage{url} 
\usepackage{booktabs}
\usepackage{amsfonts}       
\usepackage{nicefrac}       
\usepackage{microtype}      
\usepackage[noadjust]{cite}
\usepackage{array}
\usepackage{pifont}
\usepackage{capt-of,etoolbox}
\usepackage{multirow}
\usepackage{hhline}
\usepackage{bbding}
\usepackage{algorithm}
\usepackage{algorithmic}
\usepackage{boldline}
\usepackage{braket}
\usepackage{color}
\usepackage[font=footnotesize,labelfont=bf]{caption}
\usepackage{wrapfig,lipsum}
\usepackage[12pt]{moresize}
\newcommand{\cmark}{\ding{51}}%
\newcommand{\xmark}{\ding{55}}%
\usepackage[table,xcdraw]{xcolor}
\NewEnviron{NORMAL}{%
    \scalebox{0.88}{$\BODY$} 
} 

\usepackage[pagebackref=true,breaklinks=true,letterpaper=true,colorlinks=true,bookmarks=true]{hyperref}
\usepackage[capitalize]{cleveref}
\crefname{section}{Sec.}{Secs.}
\Crefname{section}{Section}{Sections}
\Crefname{table}{Table}{Tables}
\crefname{table}{Tab.}{Tabs.}

\title{\LARGE \bf Boosting Generalizability towards Zero-Shot Cross-Dataset Single-Image Indoor Depth by Meta-Initialization}

\author{Cho-Ying Wu, Yiqi Zhong, Junying Wang, and Ulrich Neumann \thanks{*Authors are with the Department of Computer Science, University of Southern California, Los Angeles, CA, USA}}

\begin{document}

\makeatletter
\let\@oldmaketitle\@maketitle
\renewcommand{\@maketitle}{\@oldmaketitle
\centering\includegraphics[width=0.70\linewidth]{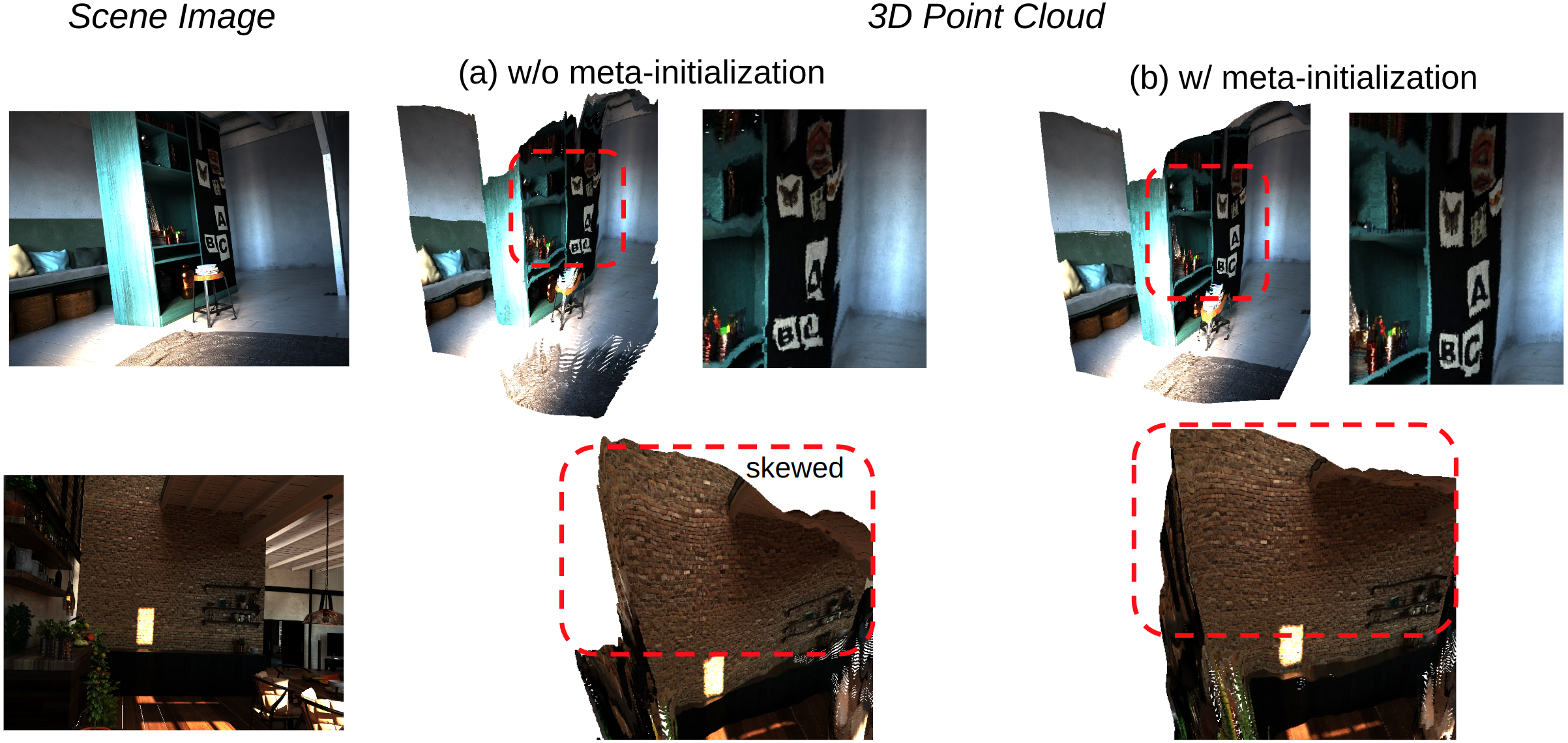}
\captionof{figure}{
\textbf{Geometry structure comparison in 3D point cloud view.} 
We back-project the predicted depth maps from images into textured 3D point cloud to show the geometry. The proposed meta-initialization has better domain generalizability that leads to more accurate depth prediction hence better 3D structures. (zoom in for the best view). 
}
\label{pointcloud}
}
\maketitle
\setcounter{figure}{1}
\begin{abstract}
Indoor robots rely on depth to perform tasks like navigation or obstacle detection, and single-image depth estimation is widely used to assist perception. Most indoor single-image depth prediction focuses less on model generalizability to unseen datasets, concerned with in-the-wild robustness for system deployment.
This work leverages gradient-based meta-learning to gain higher generalizability on zero-shot cross-dataset inference. Unlike the most-studied meta-learning of image classification associated with explicit class labels, no explicit task boundaries exist for continuous depth values tied to highly varying indoor environments regarding object arrangement and scene composition. We propose fine-grained task that treats each RGB-D mini-batch as a task in our meta-learning formulation. 
We first show that our method on limited data induces a much better prior (max 27.8\% in RMSE). Then, finetuning on meta-learned initialization consistently outperforms baselines without the meta approach. 
Aiming at generalization, we propose zero-shot cross-dataset protocols and validate higher generalizability induced by our meta-initialization, as a simple and useful plugin to many existing depth estimation methods. 
The work at the intersection of depth and meta-learning potentially drives both research to step closer to practical robotic and machine perception usage.
\end{abstract}
\section{Introduction}
\label{sec:intro}

Much research learns depth from single images to fulfill indoor robotic tasks like collision detection \cite{flacco2012depth, nascimento2020collision}, navigation \cite{tai2018socially,tan2022depth}, grasping \cite{viereck2017learning,schmidt2018grasping}, or it benefits 3D sensing or learning good 3D representations for AR/VR and view synthesis \cite{deng2022depth, roessle2022dense}. 
However, generalization is still a major issue in robustly estimating depth on unseen scenes or datasets, especially indoor scenes, since their composition varies widely, and objects are usually cluttered in the near field without an order.  
An intuitive solution is to learn from large-scale mixed datasets \cite{Ranftl2021, Ranftl2020, yin2021learning} or adopt pretrained auxiliary models as guidance \cite{wu2022toward}, but they require extra information or exogenous models. Without those resources when training on data of limited \textit{appearance and depth variation} (\textbf{scene variety} in this work), with an extreme case that only sparse and non-overlapping views are available, networks barely learn valid depth. 

Inspired by meta-learning's advantages on domain generalization and few-shot learning \cite{finn2017model, nichol2018first}, we dig into how meta-learning can be applied to pure\footnote{This differs from either stereo or video online depth adaptation \cite{zhang2019online, zhang2020online}, which assume other sequential frames exist and learn from affinity between nearby frames for online adaptation. Instead, we work on \textit{pure} singe-image problems without utilizing other frames.} single-image depth prediction.
Conventional meta-learning focuses on image classification and follows few-shot multitask learning, where a task represents a distribution to sample data from \cite{hospedales2021meta}. Instead, we study a more complex problem of scene depth estimation: the difficulties first come from estimating per-pixel and continuous range values, in contrast to global and discrete values for image classification. Next, indoor RGB-D captures vary greatly. Even within a sequence, an adjacent frame to a close view of cluttered objects can be large spaces without objects. 

This observation indicates that pure single-image depth estimation lacks clear task boundaries as conventional meta-learning~\cite{he2019task}. The problem also differs from online depth adaptation, which treats each video sequence as a task \cite{zhang2019online, zhang2020online}. To address the specific challenges, we propose to treat each mini-batch as a \textbf{fine-grained task} (Sec.~\ref{fine-grained}).


This work follows gradient-based meta-learning, which adopts a meta-optimizer and a base-optimizer \cite{finn2017model, nichol2018first}.
The base-optimizer explores multiple inner steps to find weight-update directions. 
Online and task augmentations are adopted during the base-optimization. The former augments a fine-grained task sample using color jittering and left-right flip.
The latter is after the inner steps, we further sample another fine-grained task and perform mix-up and channel shuffle.
Then, the meta-optimizer updates the meta-parameters following the inner-step explored directions.
With several epochs of dual-loops, a function $\theta^{prior}$ that maps images to depth is learned. 
$\theta^{prior}$ works as a better initialization for the subsequent supervised learning (Sec.~\ref{meta-pipeline}). Note that both meta-learning and supervised learning operate on the same training set without needing extra data. The improvements are explained via \textit{progressive learning} in Sec.~\ref{explanation}.

A previous study \cite{wu2022toward} points out that indoor depth is especially hard to robustly resolve from images due to complex object arrangements in the near field with 6-DoF camera poses that can render scenes from nearly arbitrary viewing directions. Furthermore, surface textures may cause confusion. A depth estimator needs to separate \textbf{depth-relevant/-irrelevant cues} from all extracted high-frequency features. The former indicates that depth values change when color and appearance change, such as boundaries between objects and background; the latter refers to surface textures, where depth is invariant to colors, such as paintings or material patterns on walls \cite{Chen2021S2RDepthNet,wu2022toward}.



We show that meta-learning induces a better prior with \textit{higher generalization} to unseen scenes and can identify depth-relevant/-irrelevant cues robustly.
To validate generalization, we propose evaluation protocols using multiple popular indoor datasets \cite{roberts2020hypersim, straub2019replica,ramakrishnan2021habitat,wu2022toward} for \textbf{zero-shot cross-dataset evaluation}. In contrast, most prior indoor works only train and test on a single dataset \cite{li2022depthformer, bhat2021adabins, yuan2022new}.

In experiments we show that either meta-learning on general-purpose architecture or meta-learning as plugins to several dedicated depth frameworks shows consistently superior performance. 
\textit{We highlight the finding that by only modifying training schema but keeping the same data, loss function, and architecture, performances can still be improved with more robust and generalizable depth estimation for real-world usage such as indoor robots or AR/VR surrounding perception.}
In addition to contributions in depth estimation, from meta-learning's view, we introduce fine-grained task on a continuous, per-pixel, and real-valued regression problem to advance meta-learning's study on practical problems. Fig.~\ref{pointcloud} shows sample results.

\noindent\textbf{Contributions:}
\begin{itemize}[leftmargin=*,topsep=-6pt,itemsep=-0.2ex]
  \item The first method to apply meta-learning on pure single image depth prediction, which helps achieve higher generalization in depth estimation without multiple training datasets, side information, or other pretrained networks.
  \item A novel fine-grained task concept to overcome a pure single-image setting where no explicit task boundaries exist. The work further serves as an empirical study on a practical and challenging problem in meta-learning research.
  \item A proposed protocol for zero-shot cross-dataset evaluation on indoor scenes. The protocol faithfully evaluates a model's robustness and generalizability. Extensive experiments are shown to validate our strategy.
  
\end{itemize}

\section{Related Work}
\label{related}
\noindent\textbf{Depth from Single Indoor Images}. The task has gained higher popularity \cite{li2022depthformer,li2021structdepth} while more high-quality datasets become available, such as Hypersim \cite{roberts2020hypersim}, Replica \cite{straub2019replica}, HM3D \cite{ramakrishnan2021habitat}, and VA \cite{wu2022toward}.
Some methods adopt surface normal \cite{yin2019enforcing}, plane constraints \cite{lee2019big,li2021structdepth}, advanced loss functions \cite{bhat2021adabins}, auxiliary depth completion \cite{guizilini2021sparse}, mixed-dataset training \cite{yin2021learning, Ranftl2020,Ranftl2021, zoedepth}, or modules customized for depth estimation \cite{yuan2022new,kim2022global,li2022depthformer}. In contrast, our work focuses on \textit{designing a better learning scheme} without extra data or exogenous models: we adopt the fundamental regression loss and off-the-shelf networks without loss of generalization in meta-learning methodology.
Most prior works train/test on the same dataset without validating cross-dataset performance. We instead devise zero-shot cross-dataset evaluation protocols using recent high-quality synthetic and real captures to validate generalization from meta-initialization. 

\noindent\textbf{Gradient-based Meta-Learning}. Meta-Learning principles \cite{hochreiter2001learning} illustrate an oracle about learning how to learn, especially useful on domain adaption, generalization, and few-shot learning. Popular gradient-based algorithms such as MAML \cite{finn2017model} and Reptile \cite{nichol2018first} are formulated as bilevel optimization problems using a base- and meta-optimizer. MAML uses gradients computed on the query set to update the meta-parameters. Reptile does not distinguish support and query sets and simply samples data from a task distribution for inner-loop exploration. We refer readers to \cite{hospedales2021meta} for a survey on algorithms.

The majority of meta-learning studies in vision community focuses on image \cite{qiao2020learning,chen2021meta,zhao2021learning} or pixel-level classification \cite{luo2022towards,cao2019meta,gong2021cluster}. One pioneer \cite{gao2022matters} investigates single-image pose regression that naively regresses rotation angles for one synthetic object in each image. The study may be far from real use.

Few works use meta-learning for depth but only on driving scenes with a much different problem setup
~\cite{zhang2019online,tonioni2019learning,zhang2020online, sun2022learn}. Many perform online learning and adaptation using stereo \cite{zhang2019online} or monocular videos \cite{tonioni2019learning,zhang2020online} by enforcing temporal consistency. They require affinity in nearby frames and meta-optimize within a single sequence. Our problem is arguably harder due to the pure single-image setting in highly diverse indoor structures. ~\cite{sun2022learn} works on single images but requires multiple driving datasets to build tasks and train on. We do not require multiple training sets, and we find their method is limited for indoor scenes by experiments.

\section{Methods}
\label{methods}

\begin{figure*}[bt!]
\vspace{6pt}
    \centering
    \includegraphics[width=0.75\linewidth]{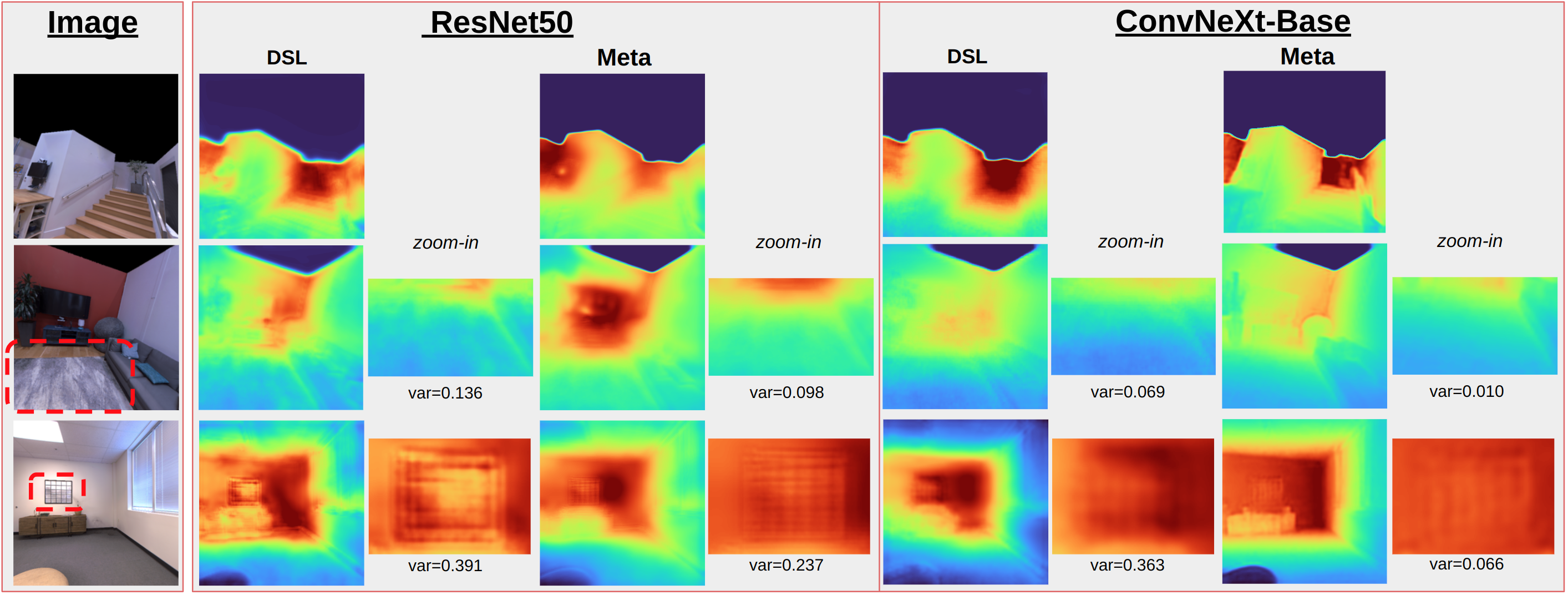}
    \vspace{-6pt}
    \caption{\textbf{Fitting to training environments.} \textit{var} shows variance for depth values in the highlighted regions. We show comparisons of fitting to training data between first-stage meta-learning (Meta) and direct supervised learning (DSL) using Replica Dataset that contains limited scene appearance and depth variation (scene variety). Meta produces smooth and more precise depth. Depth-irrelevant textures on planar regions can be resolved more correctly. In contrast, DSL produces irregularities affected by local high-frequency details, especially ResNet50. See Sec. \ref{experiments:scene-fitting} for details and \ref{explanation} for the explanation. 
    }
    \label{fitting}
    \vspace{-17pt}
\end{figure*}

\subsection{Single Mini-Batch as a Fine-Grained Task}
\vspace{-2pt}
\label{fine-grained}
\noindent\textbf{Definition}.
Single-image depth prediction learns a function $f_{\theta}:\mathcal{I}\to\mathcal{D}$, parameterized by $\theta$, to map from imagery to depth. A training set ($\mathbb{I}_{train}$, $\mathbb{D}_{train}$), containing images $I \in \mathbb{I}_{train}$ and associated depth maps $D \in \mathbb{D}_{train}$, is used to train a model. Each mini-batch with a size $K$, ($I_k$, $D_k$) $\forall k \in [1, K]$, is treated as a \textbf{fine-grained task}. 


\noindent\textbf{Difference with a task in meta-learning context}.
A fine-grained task is different from a task in the most-used meta-learning or few-shot learning context \cite{finn2017model}, where a task contains a data distribution, and batches are sampled from it. 
A fine-grained task does not contain a data distribution but is sampled from a meta-distribution, the whole RGB-D dataset.  
For example, to train a model, a navigating agent captures an RGBD dataset, which is the meta-distribution to sample each mini-batch as a fine-grained task.

\noindent\textbf{Design}. 
The design is motivated by the fact that appearance and depth variation can be arbitrarily high. 
A view looking at small desk objects and a view of large room spaces are highly dissimilar in contents and ranges, and thus mappings from their scene appearance to depth values are different. Still, they can be captured in the same environment or even in neighboring frames.
This contrasts with image classification, where class samples share a common label. 
The observation explains why we treat each mini-batch as a fine-grained task instead of each environment.


\subsection{Meta-Initialization on Depth from Single Image}
\label{meta-pipeline}
\noindent\textbf{Meta-Learning stage}.
 In the first meta-learning stage, we adopt a meta-optimizer and a base-optimizer. In each meta-iteration, a fine-grained task $B$ contains $K$ samples that are sampled from the whole training set: $(I_k, D_k)\sim (\mathbb{I}_{train}, \mathbb{D}_{train})$, $\forall k \in [1,K]$. Then we take $L$ steps to explore gradient directions that minimize the regression loss, $\mathcal{L}_{reg}$. We perform \textit{online augmentation}, $\mathbf{Aug}$, at each exploration step, including color jittering and left-right flip to craft multiple samples for a task. We get ($\theta^1_{expl}, \theta^2_{expl}, ..., \theta^L_{expl}$) from
\begin{equation}
    \vspace{-6pt}
      \NORMAL{\theta^{i}_{expl} \leftarrow \theta^{i-1}_{expl} - \alpha  \frac{1}{K}  \nabla_{\theta}\sum_{k\in [1,K]} \mathcal{L}_{reg}(\mathbf{Aug}(I_k), D_k; \theta^{i-1}_{expl}).
      }
     \vspace{-6pt}
\label{exploration}
\end{equation}

To avoid over-fitting specific to gradient-based meta-learning \cite{yao2021improving} and improve generalization, after the inner steps, we do \textit{task augmentation}, including mix-up and channel shuffle. We sample another fine-grained task $B'$ and linearly blend $B$ and $B'$ at the bottleneck $\phi=f_E(I)$ after encoder $f_E$ and interpolate the depth groundtruth.
\begin{equation}
    \vspace{-6pt}
      \NORMAL{\phi^m_k=\lambda_k f_E(I_k)+(1-\lambda_k) f_E(I'_k),\quad D^m_k=\lambda_k D_k+(1-\lambda_k) D'_k,
      }
     \vspace{-6pt}
\label{mixup}
\end{equation}
where $\forall k \in [1,K]$ and $\lambda_k \sim \mathbf{Beta}(0.5,0.5)$ following the common mixup \cite{zhang2017mixup}. $\phi^m$ is passed into the decoder head supervised by $D^m$ with loss $\mathcal{L}_{reg}$. We also perform channel shuffle at $B$'s bottleneck $\phi^B$ with a channel size of $p$ by randomly choosing a subset of channels in $\phi^B$ to be replaced by the same channels in $\phi^{B'}$. After shuffles, we get $\phi^{cs}$, which is passed into the decoder head and supervised by $B$'s depth groundtruth using the loss $\mathcal{L}_{reg}$: 
\begin{equation}
    \vspace{-6pt}
      \NORMAL{\phi^{cs}_k=r_k \phi^B +(1-r_k) \phi^{B'}, \quad r_k \sim \mathbf{Ber}(0.95),
      }
     \vspace{-6pt}
\label{cs}
\end{equation}
which means on average $p/20$ channels are shuffled with $B'$ by Bernoulli random variables, and we empirically find more shuffles can break down the representations of $B$. After inner loops and getting $\theta^L_{expl}$, either channel shuffle or mix-up is chosen with equal chances to use and optimize to get $\theta^L_{aug}$.

Subsequent to the inner steps and task augmentation, we update the meta-parameters in Reptile style~\cite{nichol2018first}, i.e., following the explored weight updates in the inner steps 
\begin{equation}
    \vspace{-6pt}
     \NORMAL{\theta^{j}_{meta} \leftarrow \theta^{j-1}_{meta} - \beta (\theta^{j-1}_{meta} - \theta^L_{aug}),
     }
     \vspace{-6pt}
\label{meta-update}
\end{equation}
where $\alpha$ and $\beta$ are respective learning rates ($lr$), and $i$ and $j$ denote inner and meta-iterations. After the last meta-iteration, we obtain meta-learned weights as the depth prior $\theta^{prior}$. 

\begin{figure}[bt!]
    \centering
    \includegraphics[width=1.0\linewidth]{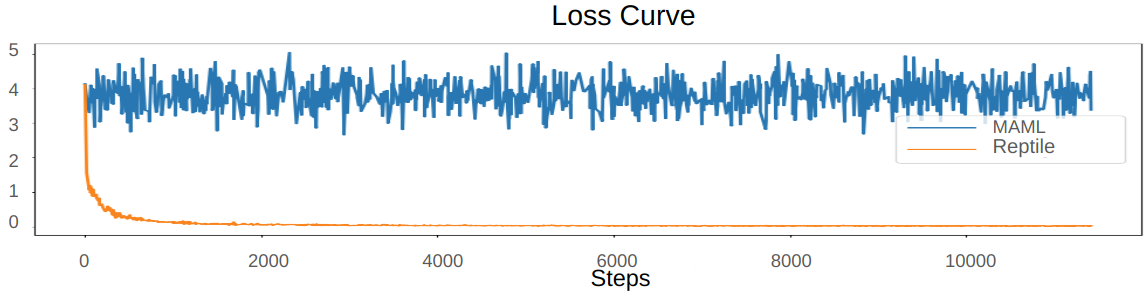}
    \vspace{-16pt}
    \caption{\textbf{Loss curve for MAML v.s. Reptile on ResNet50}. We grid search MAML's $\alpha$ and $\beta$ in [10$^{0}$, 10$^{-4}$] and cannot find parameters for convergence (a non-converging curve with $\alpha$=0.001 and $\beta$=0.5 is shown for clarity), and we plot Reptile with $\alpha$=0.001 and $\beta$=0.5 also by grid search.}
    
    \label{loss_curve}
    \vspace{-5pt}
\end{figure}

Compared with MAML \cite{finn2017model}, we find Reptile more suitable for training fine-grained tasks. In Reptile's paper, it is designed without support and query split, and thus it inherently does not require multiple samples in a task, which matches our fine-grained task definition. Next, first-order MAML computes gradients on the query set at the last inner step to update meta-parameters. Yet, without simple augmentation of color jittering and flipping, only one sample exists in each fine-grained task, and each fine-grained task differs greatly by random sampling from the dataset. Thus, if taking exploration on a support split and computing gradients on the query split, but the support and query samples are sampled from different scenes without any similarity, the gradients are nearly random and prevented from converging. This contrasts with video sequence as tasks, where they use MAML by exploiting affinity between frames~\cite{tonioni2019learning,zhang2020online}. Reptile does not require affinity between samples and naturally stabilizes training fine-grained tasks. We show the loss curves in Fig. \ref{loss_curve}.

\noindent\textbf{Supervised learning stage.}
We use $\theta^{prior}$ to initialize the subsequent supervised learning with stochastic gradient descent to minimize the regression loss, $\mathcal{L}_{reg}$, computed on the same training data ($\mathbb{I}_{train},\mathbb{D}_{train}$).
\begin{equation}
\vspace{-6pt}
     \NORMAL{\theta^* \leftarrow \min_{\theta} \mathcal{L}_{reg}(\mathbb{I}_{train},\mathbb{D}_{train}|\theta^{prior}).
     }
     \vspace{-6pt}
\label{supervised}
\end{equation}
Last, a test set ($\mathbb{I}_{test}, \mathbb{D}_{test}$) is used to evaluate the depth estimation performance of $\theta^*$.
The full procedure is shown in Algorithm \ref{meta-algo}. The implementation only needs a few lines of codes as a plugin to existing depth estimation frameworks.




\vspace{-9pt}
\begin{algorithm}
\ssmall
\caption{Meta-Initialization with fine-grained task}
\begin{algorithmic}[1]
    
    \FOR {epoch = $1:N$}
    	\FOR {$j = 1:T$ (iterations)}
    		\STATE $\theta^0_{expl}$ $\leftarrow$ $\theta^j_{meta};\ $ ($I_k$, $D_k$) $\sim$ ($\mathbb{I}_{train}$, $\mathbb{D}_{train}$), $\forall k \in [1, K]$. 
    		\FOR {$i$ = $1:L$ (steps)}
    		    \scriptsize \STATE $\theta^{i}_{expl} \leftarrow \theta^{i-1}_{expl} - \alpha  \frac{1}{K} \nabla_{\theta}\sum_{k} \mathcal{L}_{reg}(\mathbf{Aug}(I_k), D_k; \theta^{i-1}_{expl})$.
    		\ENDFOR
            \STATE Mix-up or channel shuffle to get $\theta^L_{aug}$ minimized by $\mathcal{L}_{reg}$ with $lr=\alpha$.
            \STATE $\theta^{j}_{meta} \leftarrow \theta^{j-1}_{meta} - \beta (\theta^{j-1}_{meta} - \theta^L_{aug})$.
        \ENDFOR
    \ENDFOR
    \STATE Prior $\theta^{prior} \leftarrow \theta_{meta}^T$ at epoch $N$.
    \STATE Supervised learning by $\theta^* \leftarrow \min_{\theta} \mathcal{L}_{reg}(\mathbb{I}_{train},\mathbb{D}_{train}|\theta^{prior})$.
\end{algorithmic}
\label{meta-algo}
\end{algorithm}
\vspace{-13pt}

\subsection{Strategy Explanation}
\label{explanation}
\noindent\textbf{Meta-Initialization.}
For each meta-iteration, the base-optimizer explores its neighborhood with $L$ steps.
Compared to the usual single-step update, the meta-update first takes $L$-step amortized gradient descent with a lower learning rate to delicately explore local loss manifolds. Then it updates meta-parameters by direction from the inner steps but with a step size $\beta$ towards $\theta^L$.
The meta-learned $\theta^{prior}$ may underfit a training set since it does not wholly follow optimal gradients but with a $\beta$ for control. However, it also forces the inner exploration to reach a better understanding and avoid anchoring on seen RGB-D local cues. 

\begin{figure}[bt!]
    \centering
    \vspace{4pt}
    \includegraphics[width=0.80\linewidth]{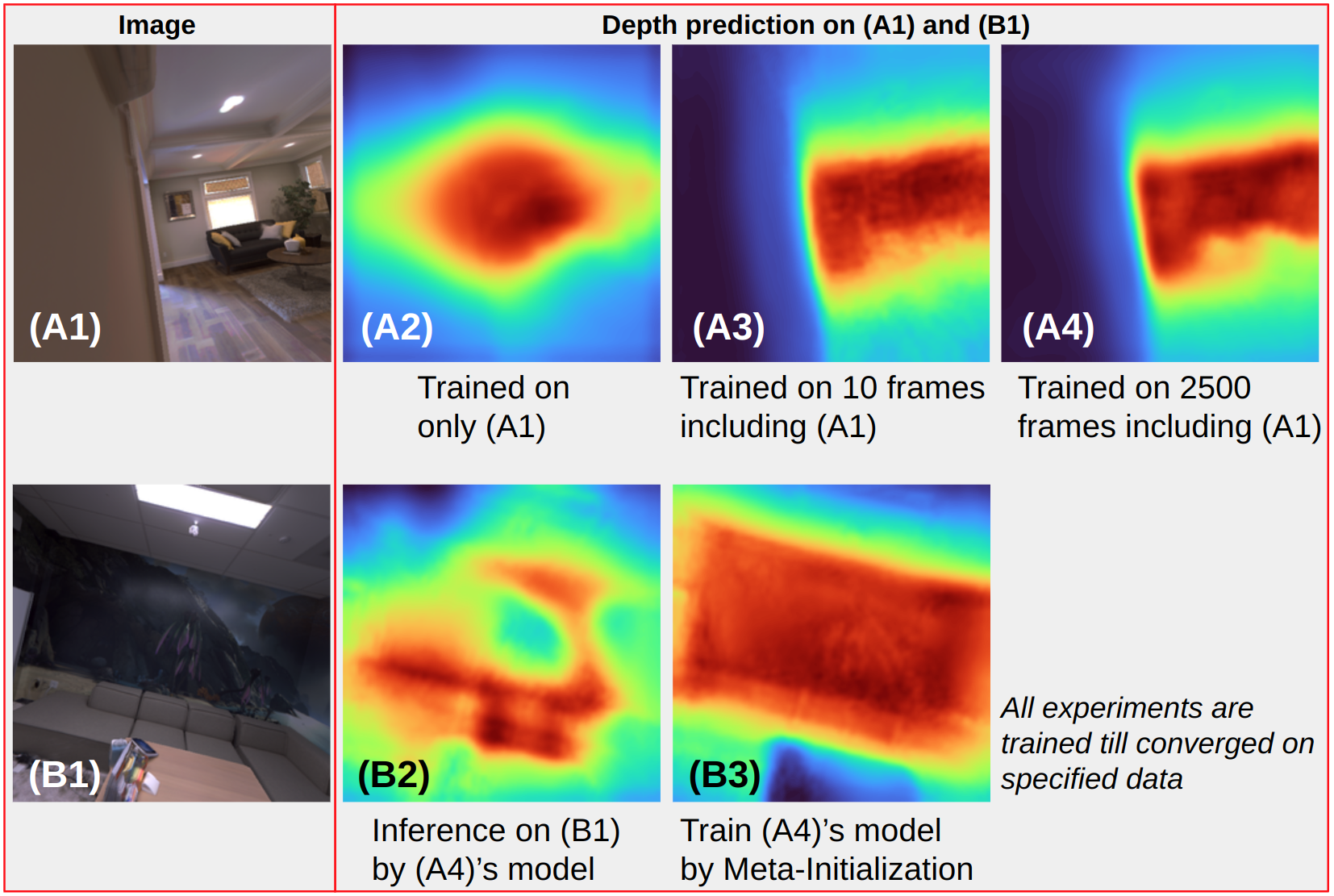}
    \vspace{-4pt}
    \caption{\textbf{Analysis on scene variety and model generalizability.} (A) shows that limited scene variety hinders learning an image-to-depth mapping, with an extreme case (A2) for only one training image. (B) shows that though a model (A4) fits well on the training scenes, it still cannot generalize to an unseen scene, especially the wall painting contains many depth-irrelevant cues. Meta-initialization attains better model generalizability (Sec.~\ref{explanation}).
    }
    \label{sv}
    \vspace{-17pt}
\end{figure}

\noindent\textbf{Progressive learning perspective.}
The above strategy can be seen as progressive learning. The first-stage meta-learning avoids anchoring on seen local cues and gets coarse but \textbf{smooth} depth.
Fig.~\ref{fitting} shows applying the first-stage meta-learning and direct supervised learning on data with limited scene variety. The meta-learning estimates smooth depth, but the direct supervised learning suffers from irregularity. 
The irregularity indicates the data cannot sufficiently demonstrate how images map to the depth domain due to the limited scene variety to learn smooth depth from global context, and thus only local high-frequency cues show up. See further analysis in Fig.~\ref{sv}.
The irregularity occurs at surface textures since those depth-irrelevant local cues are triggered but barely suppressed. 
In the second stage, a network learns finer depth based on the smoothness prior.

\section{Experiments and Discussion}
\label{experiments}
\begin{table*}[h]
\begin{center}
\vspace{4pt}
  \caption{\textbf{Cross-Dataset generalization trained by different scene varieties.} $a \to b$: training on $a$- and testing on $b$-dataset. Replica: lower scene variety; HM3D: higher scene variety. 
   Meta-Learning's gains are especially prominent in Replica$\to$VA.}
  \vspace{-5pt}
  \ssmall
  \label{table:limited}
  \begin{tabular}[c]
  {
  p{3.5cm}<{\centering\arraybackslash}|
  p{1.3cm}<{\centering\arraybackslash}|
  p{1.3cm}<{\centering\arraybackslash}|
  p{1.3cm}<{\centering\arraybackslash}|
  p{1.3cm}<{\centering\arraybackslash}|
  p{1.3cm}<{\centering\arraybackslash}|
  p{1.3cm}<{\centering\arraybackslash}|
  p{1.3cm}<{\centering\arraybackslash}|
  p{1.3cm}<{\centering\arraybackslash}}
  \hlineB{2}
  \hline
     & \multicolumn{4}{c}{\cellcolor[HTML]{99CCFF}Replica $\to$ VA} & \multicolumn{4}{c}{\cellcolor[HTML]{F7BCDC}HM3D $\to$ VA} \\
      Method & \cellcolor[HTML]{FAE5D3} MAE$\downarrow$ & \cellcolor[HTML]{FAE5D3} AbsRel$\downarrow$ & \cellcolor[HTML]{FAE5D3} RMSE$\downarrow$  & \cellcolor[HTML]{FAE5D3} SILog$\downarrow$ &  \cellcolor[HTML]{FAE5D3} MAE$\downarrow$ & \cellcolor[HTML]{FAE5D3} AbsRel$\downarrow$ & \cellcolor[HTML]{FAE5D3} RMSE$\downarrow$ & \cellcolor[HTML]{FAE5D3} SILog$\downarrow$\\
    \hline
      Direct supervised learning & 0.718 & 0.538 & 1.078 & 0.372 & 0.544 & 0.456 & 0.715 & 0.320\\
      First-Stage meta-learning (K=32) & \textbf{0.569} & \textbf{0.441} & \textbf{0.778} & \textbf{0.295} & \textbf{0.421} & \textbf{0.379} & \textbf{0.599} & \textbf{0.267}\\
      Improvement & \textit{-20.8\%} & \textit{-18.0\%} & \textit{-27.8\%} & \textit{-20.7\%} & \textit{-22.6\%} & \textit{-16.9\%} & \textit{-16.2\%} & \textit{-16.6\%} \\
    \hline
    \hlineB{2}
    First-Stage meta-learning (K=8) & 0.587 & 0.455 & 0.819 & 0.311 & 0.429 & 0.386 & 0.610 & 0.274\\
    First-Stage meta-learning (K=1) & 0.617 & 0.475 & 0.847 & 0.325 & 0.445 & 0.398 & 0.635 & 0.280\\
    \hlineB{2}
  \end{tabular}
  \vspace{-20pt}
\end{center}
\end{table*}

\noindent\textbf{Aims}. We validate our meta-initialization with four questions. \textbf{Q1} Can meta-learning improve performances on limited scene-variety datasets (Sec.~\ref{experiments:scene-fitting})? \textbf{Q2} What improvements can meta-initialization bring compared with the most popular ImageNet-initialization (Sec.~\ref{experiments:intra-dataset})? \textbf{Q3} How does meta-initialization help zero-shot cross-dataset generalization (Sec.~\ref{experiments:cross-dataset})? \textbf{Q4} How does better depth helps learning 3D representations (Sec.~\ref{experiments:nerf})?



\noindent\textbf{Datasets:} We introduce the adopted datasets as follows.
\begin{itemize}[leftmargin=*,topsep=-5pt,itemsep=0.0ex]
  \item Hypersim \cite{roberts2020hypersim} has high scene variety with 470 synthetic indoor environments, from small rooms to large open spaces, with about 67K training and 7.7K testing images.
  \item HM3D \cite{ramakrishnan2021habitat} and Replica \cite{straub2019replica} have 200K and 40K images, where we use data rendered in SimSIN \cite{wu2022toward}. HM3D has 800 scenes with high scene variety, and Replica has low scene variety with 18 overlapping scenes. 
  \item NYUv2 \cite{silberman2012indoor} contains 464 real indoor scenes with 654 testing images but with limited camera viewing direction.
  \item VA \cite{wu2022toward} as a test set has 3.5K photorealistic renderings with arbitrary camera viewing directions.
\end{itemize}
\vspace{4pt}

\noindent\textbf{Training Settings.} ResNet and ConvNeXt are backbones to extract bottleneck features. We build a depth regression head following \cite{Godard_2019_ICCV} that contains 5 convolution blocks with skip connection. Each convolution block contains a 3$\times$3 convolution, an ELU activation, and a bilinear 2$\times$ upsampling layer. Channel-size of each convolution block is $(256, 128, 64, 32, 16)$. Last, a 3$\times$3 convolution with a sigmoid activation is used to get 1-channel output depth maps. We set $N=5$, $L=4$, $K=32$, $(\alpha, \beta)=(10^{-3}, 0.5)$ for ResNet, and $(\alpha, \beta)=(5\times10^{-4}, 0.5)$ for ConvNeXt. At the supervised learning stage, we train models with a learning rate ($lr$) of $3\times10^{-4}$ till convergence. Input size to the network is 256$\times$256.  $L_2$ loss is used as $\mathcal{L}_{reg}$. 



\noindent\textbf{Metrics.} Error metrics (in meters if having physical units): Mean Absolute Error (MAE), Absolute Relative Error (AbsRel), Root Mean Square Error (RMSE), Scale Invariant Log Error (SILog). Accuracy: $\delta_C$ (percentage of correctness. Higher $\delta_C$ implies more structured and accurate depth). Correctness: the ratio of prediction and groundtruth is within $1.25^C, C=[1,2,3]$.

\subsection{Meta-Learning on Data with Limited Scene Variety} 
\label{experiments:scene-fitting}
We first examine the first-stage meta-learning on data with limited scene variety. We train $N=15$ epochs of the first-stage meta-learning with ($\alpha,\beta$) = (10$^{-4}$, 0.5) and compare with the direct supervised learning by a learning rate $lr=10^{-4}$ with an equivalent update epochs of $NL\beta=30$, where we also use the early stopping to prevent explicit overfitting. We do not use the online and task augmentation in the meta-learning. ResNet50 is used.
The other hyperparameters are the same as given in Training Settings. 
Fig.~\ref{fitting} shows fitting to training data. Replica with limited scene variety is used to verify gains in low-resource situations.
From the figure, the meta-learning is capable of identifying near/ far fields without irregularity, where the direct supervised learning struggles. Under this low-resource case, the meta-learning still induces a better image-to-depth mapping that delineates object shapes, separates depth-relevant/-irrelevant cues, and shows flat planes where rich depth-irrelevant textures exist. The observation follows the explanations in Sec.~\ref{explanation}. 

We next numerically examine generalization to unseen scenes when training on data of different-level scene variety. HM3D (high-variety) and Replica (low-variety) are used as training sets, and VA is used for testing. Table \ref{table:limited} shows that models trained by the first-stage meta-learning substantially outperform the direct supervised learning with $16.2\%$-$27.8\%$ improvements. The advantage is more evident when trained on data with low scene variety. Table~\ref{table:limited} further investigates the batch size $K$, and we find that smaller batch sizes lead to less generalization and higher variance in parameter update. Following Table~\ref{table:limited} without augmentations, we study how the augmentations improve generalization in Table~\ref{table:aug-abl}. Applying the three techniques leads to the best results that further boost the performances of $K=32$ meta-learning in Table~\ref{table:limited}.

\noindent\textbf{Comparison to other learning strategies to learn a prior}.
We first compare meta-initialization with \textit{simple pretraining with a strong weight decay} (WD-pre), which is conventionally used to smooth weight updates, and gradient accumulation (GA) with $L$ steps, which also learns weight updates from multi-step forwards. 
WD-pre baselines: We replace the first-stage meta-learning with supervised learning using a stronger weight decay (\textit{wd}) and the same learning rate as the second stage. The regular second-stage supervised learning follows. GA baseline: We replace the meta-learning dual loop with supervised learning using gradient accumulation, which accumulates gradient for $L=4$ steps and then updates the weights once, equivalent to larger-batch supervised learning to learn a prior. GA is trained with a learning rate of 0.0012, $L\times$ by the second-stage base learning rate. Then it is followed by the regular second stage. Table~\ref{table:add_pre} shows the results.
\textit{w/o prior learning} only uses the second-stage supervised learning. \textit{w/ WD-pre} (\textit{lr}=3x10$^{-4}$, \textit{wd}=10$^{-2}$) is equivalent with longer second-stage training.
The results reveal that without the interplay of dual loop optimization, simple tricks such as smoothing by weight decays or gradient accumulation do not learn better priors and show explicit gains.

\begin{figure*}[bt!]
    \centering
    \vspace{3pt}
    \includegraphics[width=0.74\linewidth]{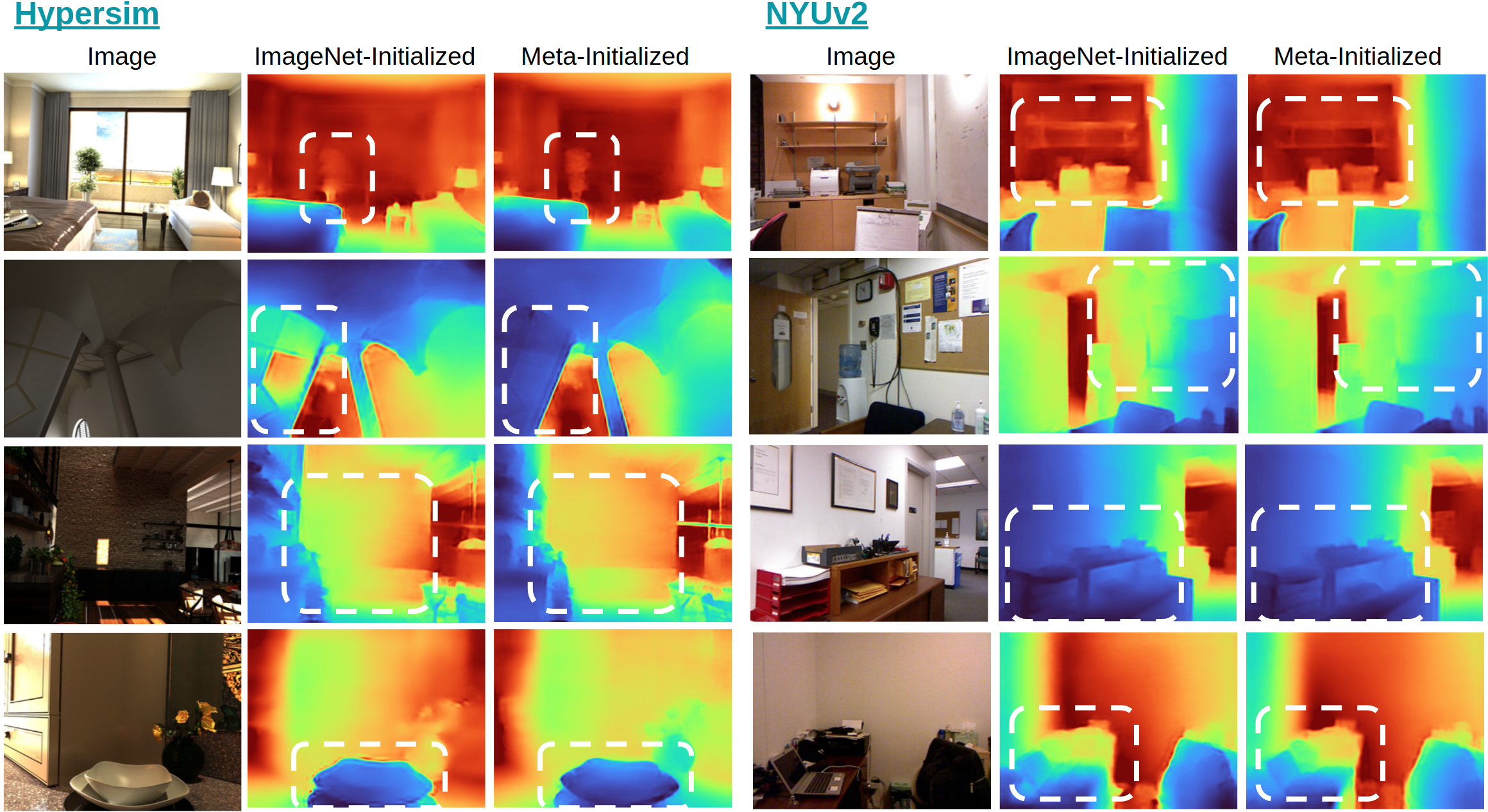}
    \vspace{-4pt}
    \caption{\textbf{Depth map qualitative comparison.} Results of our meta-initialization have better object shapes with clearer boundaries. Depth-irrelevant textures are suppressed, and flat planes are predicted, as shown in Hypersim- Row 2 ceiling and 3 textured wall examples. Zoom in for the best view.}
    \vspace{-14pt}
    \label{intra-qualitative}
\end{figure*}

\begin{table}[tb!]
\begin{center}
  \caption{\textbf{Augmentation studies} including Online Augmentation (OnAug), mix-up (MX), and channel shuffle (CS). Discarding them all means not using augmentations in the first stage. Setting: Replica$\to$VA following Table~\ref{table:limited}.
  }
  \vspace{-4pt}
  \ssmall
  \label{table:aug-abl}
  \begin{tabular}[c]
  {
  p{0.25cm}<{\centering\arraybackslash}
  p{0.25cm}<{\centering\arraybackslash}
  p{0.25cm}<{\centering\arraybackslash}
  p{0.60cm}<{\centering\arraybackslash}
  p{0.60cm}<{\centering\arraybackslash}
  p{0.60cm}<{\centering\arraybackslash}
  p{0.55cm}<{\centering\arraybackslash}}
  \hlineB{2}
      \multicolumn{1}{|c|}{\cellcolor[HTML]{F1F8B5} OnAug} & \multicolumn{1}{|c|}{\cellcolor[HTML]{F1F8B5} MX} & \multicolumn{1}{|c|}{\cellcolor[HTML]{F1F8B5} CS} & \cellcolor[HTML]{FAE5D3} MAE$\downarrow$ & \cellcolor[HTML]{FAE5D3} AbsRel$\downarrow$ & \cellcolor[HTML]{FAE5D3} RMSE$\downarrow$  &  \multicolumn{1}{|c|}{\cellcolor[HTML]{FAE5D3} SILog$\downarrow$}  \\
    \hline
    \xmark & \xmark & \xmark & 0.569 & 0.441 & 0.778 & 0.295 \\
    \cmark & \xmark & \xmark & 0.548 & 0.430 & 0.761 & 0.288 \\
    \cmark & \cmark & \xmark & 0.528 & 0.425 & 0.740 & 0.283 \\
    \cmark & \xmark & \cmark & 0.531 & 0.427 & 0.748 & 0.285 \\
    \xmark & \cmark & \cmark & 0.524 & 0.429 & 0.744 & 0.284 \\
    \cmark & \cmark & \cmark & 0.515 & 0.418 & 0.729 & 0.282 \\

    \hlineB{2}
    \hline
  \end{tabular}

  \vspace{-12pt}
\end{center}
\end{table}

\begin{table}[bt!]
\begin{center}
  \caption{\textbf{Meta-Initialization v.s. simple learning strategies.} Stronger weight decay pretraining (WD-pre) and gradient accumulation (GA) are compared against. We use ResNet50 and train/ test on NYUv2.}
  \vspace{-8pt}
  \ssmall
  \label{table:add_pre}
  \begin{tabular}[c]
  {
  p{2.3cm}<{\arraybackslash}
  p{0.65cm}<{\centering\arraybackslash}
  p{0.60cm}<{\centering\arraybackslash}
  p{0.55cm}<{\centering\arraybackslash}
  p{0.25cm}<{\centering\arraybackslash}
  p{0.25cm}<{\centering\arraybackslash}
  p{0.25cm}<{\centering\arraybackslash}}
  \hlineB{2}
  
      \multicolumn{1}{|c|}{\cellcolor[HTML]{99CCFF} NYUv2}  & \cellcolor[HTML]{FAE5D3} AbsRel$\downarrow$ & \cellcolor[HTML]{FAE5D3} RMSE$\downarrow$ & \cellcolor[HTML]{FAE5D3} SILog$\downarrow$ & \multicolumn{1}{|c|}{\cellcolor[HTML]{D5F5E3} $\delta_1$$\uparrow$} &  \multicolumn{1}{|c|}{\cellcolor[HTML]{D5F5E3} $\delta_2$$\uparrow$} &  \multicolumn{1}{|c|}{\cellcolor[HTML]{D5F5E3} $\delta_3$$\uparrow$} \\
      \hline
      w/o prior learning & 0.131 & 0.480 & 0.175 & 83.6 & 96.4 & 99.0 \\
      w/ WD-pre (\textit{wd}=0.1) & 0.133 & 0.484 & 0.176 & 83.7 & 96.6 & 98.9 \\
      w/ WD-pre (\textit{wd}=0.05) & 0.130 & 0.479 & 0.174 & 83.6 & 96.3 & 98.8  \\
      w/ WD-pre (\textit{wd}=0.01) & 0.130 & 0.478 & 0.173 & 83.6 & 96.4 & 99.0 \\
      w/ GA & 0.133 & 0.484 & 0.175 & 83.8 & 96.4 & 98.9 \\
      w/ Meta-Initialization & \textbf{0.122} & \textbf{0.454}  & \textbf{0.167} & \textbf{85.4} & \textbf{96.8} & \textbf{99.3} \\
    \hlineB{2}
    \hline
  \end{tabular}
  \vspace{-15pt}
\end{center}
\end{table}

\begin{table}[tb!]
\label{table:intra}
\begin{center}
  \caption{\textbf{Intra-Dataset evaluation.} We evaluate full Algorithm \ref{meta-algo} (meta-initialization, "+Meta") on NYUv2 and Hypersim and train/test on the same dataset. +Meta consistently outperforms ImageNet-initialization (no marks). 
  }
  \vspace{-8pt}
  \ssmall
  \label{table:intra-dataset}
  \begin{tabular}[c]
  {
  p{2.25cm}<{\arraybackslash}
  p{0.65cm}<{\centering\arraybackslash}
  p{0.60cm}<{\centering\arraybackslash}
  p{0.55cm}<{\centering\arraybackslash}
  p{0.25cm}<{\centering\arraybackslash}
  p{0.25cm}<{\centering\arraybackslash}
  p{0.25cm}<{\centering\arraybackslash}}
  \hlineB{2}
  
      \multicolumn{1}{|c|}{\cellcolor[HTML]{99CCFF} Hypersim}  & \cellcolor[HTML]{FAE5D3} AbsRel$\downarrow$ & \cellcolor[HTML]{FAE5D3} RMSE$\downarrow$ & \cellcolor[HTML]{FAE5D3} SILog$\downarrow$ & \multicolumn{1}{|c|}{\cellcolor[HTML]{D5F5E3} $\delta_1$$\uparrow$} &  \multicolumn{1}{|c|}{\cellcolor[HTML]{D5F5E3} $\delta_2$$\uparrow$} &  \multicolumn{1}{|c|}{\cellcolor[HTML]{D5F5E3} $\delta_3$$\uparrow$} \\
    \hline
      ResNet50     & 0.248 & 1.775 & 0.261 & 64.8 & 87.1 & 94.7 \\
      ResNet50+Meta & \textbf{0.235} & \textbf{1.659} & \textbf{0.246} & \textbf{66.9} & \textbf{88.0} & \textbf{95.1} \\
      \hline
      ResNet101    & 0.234 & 1.671 & 0.243 & 67.4 & 88.5 & 95.3 \\
      ResNet101+Meta   & \textbf{0.215} & \textbf{1.563} & \textbf{0.236} &  \textbf{68.2} & \textbf{89.1} & \textbf{95.4}\\
      \hline
      ConvNeXt-base   & 0.201 & 1.534 & 0.221 & 73.6 & 91.1 & 96.3 \\
      ConvNeXt-base+Meta  & \textbf{0.184} & \textbf{1.411} & \textbf{0.210} & \textbf{75.0} & \textbf{91.8} & \textbf{96.5} \\ 
    \hlineB{2}
    \hline
  \end{tabular}

  \begin{tabular}[c]
  {
  p{2.25cm}<{\arraybackslash}
  p{0.65cm}<{\centering\arraybackslash}
  p{0.60cm}<{\centering\arraybackslash}
  p{0.55cm}<{\centering\arraybackslash}
  p{0.25cm}<{\centering\arraybackslash}
  p{0.25cm}<{\centering\arraybackslash}
  p{0.25cm}<{\centering\arraybackslash}}
  \hlineB{2}
  
     \multicolumn{1}{|c|}{\cellcolor[HTML]{99CCFF} NYUv2}  & \cellcolor[HTML]{FAE5D3} AbsRel$\downarrow$ & \cellcolor[HTML]{FAE5D3} RMSE$\downarrow$ & \cellcolor[HTML]{FAE5D3} SILog$\downarrow$ & \multicolumn{1}{|c|}{\cellcolor[HTML]{D5F5E3} $\delta_1$$\uparrow$} &  \multicolumn{1}{|c|}{\cellcolor[HTML]{D5F5E3} $\delta_2$$\uparrow$} &  \multicolumn{1}{|c|}{\cellcolor[HTML]{D5F5E3} $\delta_3$$\uparrow$} \\
    \hline
      ResNet50    & 0.131 & 0.480 & 0.175 & 83.6 & 96.4 & \textbf{99.0} \\
      ResNet50+Meta   & \textbf{0.120} & \textbf{0.448} & \textbf{0.167} &\textbf{85.5} & \textbf{96.8} & \textbf{99.0} \\
      \hline
      ResNet101    & 0.120 & 0.448 & 0.167 & 85.6 & 97.1 & \textbf{99.3} \\
      ResNet101+Meta   & \textbf{0.109} & \textbf{0.416} & \textbf{0.158} & \textbf{86.8} & \textbf{97.2} & \textbf{99.3} \\
      \hline
      ConvNeXt-base   & 0.101 & 0.394 & 0.138& 89.4 & 97.9 & \textbf{99.5} \\ 
      ConvNeXt-base+Meta   & \textbf{0.097} & \textbf{0.385} & \textbf{0.133} & \textbf{89.8} & \textbf{98.2} & \textbf{99.5} \\ 
      \hline
    \hlineB{2}
    \hline
  \end{tabular}
  \vspace{-23pt}
\end{center}
\end{table}

\begin{table}[tb]
\begin{center}
\vspace{3pt}
  \caption{\textbf{Cross-Dataset evaluation on general architecture.} Comparison: ImageNet-initialization (no marks) and meta-initialization (+Meta). 
  }
  \vspace{-4pt}
  \ssmall
  \label{table:cross-dataset-hm3d}
  
  \begin{tabular}[c]
  {
  p{2.39cm}<{\arraybackslash}
  p{0.65cm}<{\centering\arraybackslash}
  p{0.60cm}<{\centering\arraybackslash}
  p{0.55cm}<{\centering\arraybackslash}
  p{0.25cm}<{\centering\arraybackslash}
  p{0.25cm}<{\centering\arraybackslash}
  p{0.25cm}<{\centering\arraybackslash}}
  \hlineB{2}
  
       \multicolumn{1}{|c|}{\cellcolor[HTML]{99CCFF} HM3D $\to$ VA}  & \cellcolor[HTML]{FAE5D3} AbsRel$\downarrow$ & \cellcolor[HTML]{FAE5D3} RMSE$\downarrow$ & \cellcolor[HTML]{FAE5D3} SILog$\downarrow$ & \multicolumn{1}{|c|}{\cellcolor[HTML]{D5F5E3} $\delta_1$$\uparrow$} &  \multicolumn{1}{|c|}{\cellcolor[HTML]{D5F5E3} $\delta_2$$\uparrow$} &  \multicolumn{1}{|c|}{\cellcolor[HTML]{D5F5E3} $\delta_3$$\uparrow$} \\
    \hline
      ConvNeXt-small & 0.180 & 0.389 & 0.166 & 74.6 & 91.0 & 96.1 \\
      ConvNeXt-small+Meta & \textbf{0.159} & \textbf{0.340} & \textbf{0.145} & \textbf{78.0} & \textbf{93.2} & \textbf{97.3} \\
      ConvNeXt-base  & 0.176 & 0.385 & 0.155 & 76.1 & 91.1 & 95.4 \\ 
      ConvNeXt-base+Meta  & \textbf{0.161} & \textbf{0.352} & \textbf{0.144} & \textbf{78.1} & \textbf{92.6} & \textbf{96.7} \\
      ConvNeXt-large  & 0.170 & 0.357 & 0.147 & 78.1 & 91.5 & 95.7 \\
      ConvNeXt-large+Meta & \textbf{0.156} & \textbf{0.322} & \textbf{0.142} & \textbf{79.0} & \textbf{92.3} & \textbf{96.4} \\
    \hlineB{2}
    
  
       \multicolumn{1}{|c|}{\cellcolor[HTML]{99CCFF} HM3D $\to$ NYUv2}  & \cellcolor[HTML]{FAE5D3} AbsRel$\downarrow$ & \cellcolor[HTML]{FAE5D3} RMSE$\downarrow$ & \cellcolor[HTML]{FAE5D3} SILog$\downarrow$ & \multicolumn{1}{|c|}{\cellcolor[HTML]{D5F5E3} $\delta_1$$\uparrow$} &  \multicolumn{1}{|c|}{\cellcolor[HTML]{D5F5E3} $\delta_2$$\uparrow$} &  \multicolumn{1}{|c|}{\cellcolor[HTML]{D5F5E3} $\delta_3$$\uparrow$} \\
    \hline
      ConvNeXt-small  & 0.213 & 0.728 & 0.255 & 69.2 & 88.7 & 95.8 \\
      ConvNeXt-small+Meta & \textbf{0.202} & \textbf{0.706} & \textbf{0.242} & \textbf{70.8} & \textbf{89.1} & \textbf{96.0} \\
      ConvNeXt-base  & 0.208 & 0.717 & 0.252 & 70.1 & 89.4 & 96.0 \\
      ConvNeXt-base+Meta & \textbf{0.197} & \textbf{0.686} & \textbf{0.246} & \textbf{71.5} & \textbf{89.8} & \textbf{96.2} \\
      ConvNeXt-large & 0.192 & 0.690 & 0.244 & 72.0 & 90.4 & 96.4 \\ 
      ConvNeXt-large+Meta & \textbf{0.186} & \textbf{0.654} & \textbf{0.239} & \textbf{73.3} & \textbf{90.7} & \textbf{96.6} \\ 
    \hlineB{2}
    
    \multicolumn{1}{|c|}{\cellcolor[HTML]{99CCFF} HM3D $\to$ Replica}  & \cellcolor[HTML]{FAE5D3} AbsRel$\downarrow$ & \cellcolor[HTML]{FAE5D3} RMSE$\downarrow$ & \cellcolor[HTML]{FAE5D3} SILog$\downarrow$ & \multicolumn{1}{|c|}{\cellcolor[HTML]{D5F5E3} $\delta_1$$\uparrow$} &  \multicolumn{1}{|c|}{\cellcolor[HTML]{D5F5E3} $\delta_2$$\uparrow$} &  \multicolumn{1}{|c|}{\cellcolor[HTML]{D5F5E3} $\delta_3$$\uparrow$} \\
    \hline
      ConvNeXt-small & 0.138 & 0.321 & 0.095 & 84.5 & 93.9 & 96.6 \\
      ConvNeXt-small+Meta & \textbf{0.124} & \textbf{0.282} & \textbf{0.089} & \textbf{85.7} & \textbf{95.7} & \textbf{98.0} \\
      ConvNeXt-base & 0.134 & 0.316 & 0.097 & 84.6 & 94.2 & 96.6 \\ 
      ConvNeXt-base+Meta & \textbf{0.116} & \textbf{0.275} & \textbf{0.086} & \textbf{87.2} & \textbf{96.5} & \textbf{98.5} \\
      ConvNeXt-large & 0.137 & 0.307 & 0.099 & 84.3 & 94.0 & 96.6 \\
      ConvNeXt-large+Meta  & \textbf{0.115} & \textbf{0.270} & \textbf{0.084} & \textbf{87.2} & \textbf{96.5} & \textbf{98.7} \\
    \hlineB{2}
    \hline
  
       \multicolumn{1}{|c|}{\cellcolor[HTML]{99CCFF} Hypersim $\to$ VA}  & \cellcolor[HTML]{FAE5D3} AbsRel$\downarrow$ & \cellcolor[HTML]{FAE5D3} RMSE$\downarrow$ & \cellcolor[HTML]{FAE5D3} SILog$\downarrow$ & \multicolumn{1}{|c|}{\cellcolor[HTML]{D5F5E3} $\delta_1$$\uparrow$} &  \multicolumn{1}{|c|}{\cellcolor[HTML]{D5F5E3} $\delta_2$$\uparrow$} &  \multicolumn{1}{|c|}{\cellcolor[HTML]{D5F5E3} $\delta_3$$\uparrow$} \\
    \hline
      ConvNeXt-small & 0.215 & 0.404 & 0.205 & 68.5 & 90.8 & 96.7 \\
      ConvNeXt-small+Meta & \textbf{0.207} & \textbf{0.398} & \textbf{0.196} & \textbf{70.4} & \textbf{91.3} & \textbf{97.0} \\
      ConvNeXt-base  & 0.201 & 0.393 & 0.188 & 71.3 & 91.8 & 97.3 \\
      ConvNeXt-base+Meta & \textbf{0.194} & \textbf{0.365} & \textbf{0.173} & \textbf{72.8} & \textbf{92.8} & \textbf{97.8} \\ 
      ConvNeXt-large  & 0.198 & 0.369 & 0.175 & 73.0 & 92.0 & 97.1 \\
      ConvNeXt-large+Meta  & \textbf{0.183} & \textbf{0.355} & \textbf{0.164} & \textbf{74.6} & \textbf{93.5} & \textbf{97.8} \\
    \hlineB{2}
    
    \multicolumn{1}{|c|}{\cellcolor[HTML]{99CCFF} Hypersim $\to$ NYUv2} & \cellcolor[HTML]{FAE5D3} AbsRel$\downarrow$ & \cellcolor[HTML]{FAE5D3} RMSE$\downarrow$ & \cellcolor[HTML]{FAE5D3} SILog$\downarrow$ & \multicolumn{1}{|c|}{\cellcolor[HTML]{D5F5E3} $\delta_1$$\uparrow$} &  \multicolumn{1}{|c|}{\cellcolor[HTML]{D5F5E3} $\delta_2$$\uparrow$} &  \multicolumn{1}{|c|}{\cellcolor[HTML]{D5F5E3} $\delta_3$$\uparrow$} \\
    \hline
      ConvNeXt-small & 0.165 & 0.598 & 0.225 & 75.7 & 94.3 & 98.5 \\
      ConvNeXt-small+Meta & \textbf{0.155} & \textbf{0.575} & \textbf{0.208} &\textbf{77.8} & \textbf{95.1} & \textbf{98.8} \\
      ConvNeXt-base & 0.150 & 0.549 & 0.200 & 79.6 & 95.6 & 98.9 \\
      ConvNeXt-base+Meta & \textbf{0.141} & \textbf{0.524} & \textbf{0.192} & \textbf{80.3} & \textbf{96.0} & \textbf{99.0} \\
      ConvNeXt-large & 0.149 & 0.542 & 0.199 & 79.8 & 95.6 & 98.8 \\
      ConvNeXt-large+Meta & \textbf{0.140} & \textbf{0.517} & \textbf{0.187} & \textbf{81.2} & \textbf{96.2} & \textbf{99.1} \\
    \hline
    \multicolumn{1}{|c|}{\cellcolor[HTML]{99CCFF} Hypersim $\to$ Replica} & \cellcolor[HTML]{FAE5D3} AbsRel$\downarrow$ & \cellcolor[HTML]{FAE5D3} RMSE$\downarrow$ & \cellcolor[HTML]{FAE5D3} SILog$\downarrow$ & \multicolumn{1}{|c|}{\cellcolor[HTML]{D5F5E3} $\delta_1$$\uparrow$} &  \multicolumn{1}{|c|}{\cellcolor[HTML]{D5F5E3} $\delta_2$$\uparrow$} &  \multicolumn{1}{|c|}{\cellcolor[HTML]{D5F5E3} $\delta_3$$\uparrow$} \\
    \hline
      ConvNeXt-small & 0.189 & 0.417 & 0.169 & 72.4 & 92.1 & \textbf{97.5} \\
      ConvNeXt-small+Meta & \textbf{0.178} & \textbf{0.404} & \textbf{0.150} & \textbf{74.5} & \textbf{92.7} & \textbf{97.5} \\
      ConvNeXt-base  & 0.185 & 0.409 & 0.154 & 74.1 & 92.6 & 97.4 \\ 
      ConvNeXt-base+Meta & \textbf{0.173} & \textbf{0.399} & \textbf{0.142} & \textbf{75.6} & \textbf{93.3} & \textbf{97.9} \\ 
      ConvNeXt-large & 0.172 & 0.394 & 0.145 & 75.8 & 93.2 & 97.7 \\
      ConvNeXt-large+Meta & \textbf{0.165} & \textbf{0.380} & \textbf{0.132} & \textbf{77.0} & \textbf{94.0} & \textbf{98.1} \\
    \hlineB{2}
    \hline
  \end{tabular}
  \vspace{-12pt}
\end{center}
\end{table}

\begin{table}[htb!]
\begin{center}
  \caption{\textbf{Cross-Dataset evaluation using dedicated depth estimation networks.} Our meta-initialization (+Meta) can be plugged into several methods to stably improve them.}
  \vspace{-4pt}
  \ssmall
  \label{table:cross-dataset-dedicated}
  
  \begin{tabular}[c]
  {
  p{2.39cm}<{\arraybackslash}
  p{0.65cm}<{\centering\arraybackslash}
  p{0.60cm}<{\centering\arraybackslash}
  p{0.55cm}<{\centering\arraybackslash}
  p{0.25cm}<{\centering\arraybackslash}
  p{0.25cm}<{\centering\arraybackslash}
  p{0.25cm}<{\centering\arraybackslash}}
  \hlineB{2}
  
       \multicolumn{1}{|c|}{\cellcolor[HTML]{99CCFF} Hypersim $\to$ Replica} & \cellcolor[HTML]{FAE5D3} AbsRel$\downarrow$ & \cellcolor[HTML]{FAE5D3} RMSE$\downarrow$ & \cellcolor[HTML]{FAE5D3} SILog$\downarrow$ & \multicolumn{1}{|c|}{\cellcolor[HTML]{D5F5E3} $\delta_1$$\uparrow$} &  \multicolumn{1}{|c|}{\cellcolor[HTML]{D5F5E3} $\delta_2$$\uparrow$} &  \multicolumn{1}{|c|}{\cellcolor[HTML]{D5F5E3} $\delta_3$$\uparrow$} \\
    \hline
    BTS-ResNet101 \cite{lee2019big} & 0.214 & 0.488 & 0.211 & 69.9 & 89.8& 96.3\\
    BTS-ResNet101+Meta & \textbf{0.194} & \textbf{0.463} & \textbf{0.177} & \textbf{71.2} & \textbf{90.5}& \textbf{96.8}\\
    \hline
    DepthFormer \cite{li2022depthformer} & 0.185 & 0.415 & 0.166 & 72.9 & 92.3 & 97.4 \\
      DepthFormer+Meta  & \textbf{0.171} & \textbf{0.394} & \textbf{0.144} & \textbf{74.5} & \textbf{92.8} & \textbf{97.6} \\
      \hline
      DPT-hybrid \cite{Ranftl2021} & 0.197 & 0.455 & 0.175 & 71.2 & 90.7 & 96.7 \\ 
      DPT-hybrid+Meta & \textbf{0.165} & \textbf{0.391} & \textbf{0.154} & \textbf{75.7} & \textbf{93.5} & \textbf{97.8} \\
      DPT-large \cite{Ranftl2021} & 0.172 & 0.401 & 0.158 & 75.4 & 93.3 & 97.6 \\
      DPT-large+Meta & \textbf{0.161} & \textbf{0.374} & \textbf{0.134} & \textbf{77.4} & \textbf{94.4} & \textbf{98.2} \\
    \hline
    AdaBins \cite{bhat2021adabins} & 0.210 & 0.445 & 0.195 & 70.2 & 90.1 & 96.6 \\
    AdaBins+Meta  & \textbf{0.193} & \textbf{0.427} & \textbf{0.177} & \textbf{72.0} & \textbf{92.1} & \textbf{97.4}\\
    \hline
    GLPDepth \cite{kim2022global} & 0.188 & 0.418 & 0.165 & 72.9 & 92.4& 97.6\\
    GLPDepth+Meta & \textbf{0.170} & \textbf{0.396} & \textbf{0.149} & \textbf{74.8} & \textbf{93.1}& \textbf{97.7}\\
    \hline
    NDDepth~\cite{shao2023nddepth} & 0.165 & 0.380 & 0.144 & 76.1 & 93.5 & \textbf{98.0} \\
    NDDepth+Meta & \textbf{0.159} & \textbf{0.371} & \textbf{0.136} & \textbf{76.7} & \textbf{93.8} & 97.9\\
    \hlineB{2}
    \hline
  
   \multicolumn{1}{|c|}{\cellcolor[HTML]{99CCFF} Hypersim $\to$ NYUv2} & \cellcolor[HTML]{FAE5D3} AbsRel$\downarrow$ & \cellcolor[HTML]{FAE5D3} RMSE$\downarrow$ & \cellcolor[HTML]{FAE5D3} SILog$\downarrow$ & \multicolumn{1}{|c|}{\cellcolor[HTML]{D5F5E3} $\delta_1$$\uparrow$} &  \multicolumn{1}{|c|}{\cellcolor[HTML]{D5F5E3} $\delta_2$$\uparrow$} &  \multicolumn{1}{|c|}{\cellcolor[HTML]{D5F5E3} $\delta_3$$\uparrow$} \\
    \hline
    BTS-ResNet101 \cite{lee2019big} & 0.187 & 0.641 & 0.243 & 72.3 & 90.8 & 95.8 \\
    BTS-ResNet101+Meta  & \textbf{0.170} & \textbf{0.618} & \textbf{0.226} &\textbf{74.5} & \textbf{92.8} & \textbf{97.5}\\
    \hline
      DepthFormer \cite{li2022depthformer} & 0.169 & 0.608 & 0.225 & 75.1 & 93.9 & 98.2 \\
      DepthFormer+Meta  & \textbf{0.151} & \textbf{0.572} & \textbf{0.204} & \textbf{78.1} & \textbf{94.3} & \textbf{98.3} \\
    \hline
      DPT-hybrid \cite{Ranftl2021} & 0.149 & 0.580 & 0.204 & 78.9 & 94.8 & 98.3 \\ 
      DPT-hybrid+Meta & \textbf{0.135} & \textbf{0.550} & \textbf{0.182} & \textbf{81.5} & \textbf{96.6} & \textbf{99.1} \\
      DPT-large \cite{Ranftl2021} & 0.136 & 0.530 & 0.180 & 82.3 & 96.2 & 98.8 \\
      DPT-large+Meta & \textbf{0.130} & \textbf{0.507} & \textbf{0.176} & \textbf{83.1} & \textbf{96.6} & \textbf{99.0} \\
    \hline
    AdaBins \cite{bhat2021adabins} & 0.188 & 0.642 & 0.241 & 72.6 &91.2 &96.6 \\
    AdaBins+Meta  & \textbf{0.174} & \textbf{0.622} & \textbf{0.231} & \textbf{74.1} & \textbf{92.7} & \textbf{97.5}\\
    \hline
      GLPDepth \cite{kim2022global} & 0.169 & 0.604 & 0.223 & 75.3 & 93.9 & 98.2 \\
      GLPDepth+Meta  & \textbf{0.155} & \textbf{0.577} & \textbf{0.208} & \textbf{77.8} & \textbf{94.2} & \textbf{98.4} \\
    \hline
    NDDepth~\cite{shao2023nddepth} & 0.141 & 0.557 & 0.195 & 80.4 & 96.0 & \textbf{98.8} \\
    NDDepth+Meta & \textbf{0.134} & \textbf{0.545} & \textbf{0.184} & \textbf{81.3} & \textbf{96.3} & \textbf{98.8}\\
    \hlineB{2}
    \hline
  \end{tabular}
  \vspace{-18pt}
\end{center}
\end{table}

\begin{table}[h]
\begin{center}
\vspace{3pt}
  \caption{\textbf{Comparison to meta-learning on single-image depth.} We aggregate Hypersim, HM3D, and NYUv2 as the training sets and test on VA.}
  \vspace{-5pt}
  \ssmall
  \label{table:learnAdapt}
  \begin{tabular}[c]
  {
  p{2.5cm}<{\arraybackslash}|
  p{0.8cm}<{\centering\arraybackslash}|
  p{0.8cm}<{\centering\arraybackslash}|
  p{0.6cm}<{\centering\arraybackslash}|
  p{0.6cm}<{\centering\arraybackslash}|
  p{0.6cm}<{\centering\arraybackslash}}
  \hlineB{2}
  
      \multicolumn{1}{|c|}{\cellcolor[HTML]{99CCFF} Multiple$\to$VA}  &  \cellcolor[HTML]{FAE5D3} AbsRel$\downarrow$ & \cellcolor[HTML]{FAE5D3} RMSE$\downarrow$ &  \multicolumn{1}{|c|}{\cellcolor[HTML]{D5F5E3} $\delta_1$$\uparrow$} &  \multicolumn{1}{|c|}{\cellcolor[HTML]{D5F5E3} $\delta_2$$\uparrow$} &  \multicolumn{1}{|c|}{\cellcolor[HTML]{D5F5E3} $\delta_3$$\uparrow$} \\
    \hline
    Learn to adapt ~\cite{sun2022learn} & 0.318 & 0.133 & 79.5 & 92.4 & 96.4 \\
    Our Meta-Initialization & \textbf{0.280} & \textbf{0.109} & \textbf{82.2} & \textbf{93.4} & \textbf{97.4} \\
      \hline
    \hlineB{2}
  \end{tabular}
  \vspace{-18pt}
\end{center}
\end{table}

\begin{table}[h]
\begin{center}
\vspace{7pt}
  \caption{\textbf{Cross-Dataset comparison to domain adaptation.} We follow the settings from S2R-DepthNet~\cite{Chen2021S2RDepthNet} and compare with their reported benchmark.}
  \vspace{-5pt}
  \ssmall
  \label{table:add1}
  \begin{tabular}[c]
  {
  p{2.7cm}<{\arraybackslash}|
  p{0.8cm}<{\centering\arraybackslash}|
  p{0.8cm}<{\centering\arraybackslash}|
  p{0.45cm}<{\centering\arraybackslash}|
  p{0.45cm}<{\centering\arraybackslash}|
  p{0.45cm}<{\centering\arraybackslash}}
  \hlineB{2}
  
      \multicolumn{1}{|c|}{\cellcolor[HTML]{99CCFF} SUNCG$\to$NYUv2}  &  \cellcolor[HTML]{FAE5D3} AbsRel$\downarrow$ & \cellcolor[HTML]{FAE5D3} RMSE$\downarrow$ &  \multicolumn{1}{|c|}{\cellcolor[HTML]{D5F5E3} $\delta_1$$\uparrow$} &  \multicolumn{1}{|c|}{\cellcolor[HTML]{D5F5E3} $\delta_2$$\uparrow$} &  \multicolumn{1}{|c|}{\cellcolor[HTML]{D5F5E3} $\delta_3$$\uparrow$} \\
    \hline
    T$^2$Net ~\cite{zheng2018t2net} & 0.203 & 0.738 & 67.0 & 89.1 & 96.6 \\
    ARC~\cite{zhao2020domain} & 0.186 & 0.710 & 71.2 & 91.7 & 97.7\\
      S2R-DepthNet ~\cite{Chen2021S2RDepthNet} & 0.196 & 0.662 & 69.5 & 91.0 & 97.2 \\
      Our Meta-Initialization & \textbf{0.177} & \textbf{0.635} & \textbf{72.8} & \textbf{92.8} & \textbf{97.8} \\
      \hline
    \hlineB{2}
  \end{tabular}
  \vspace{-5pt}
\end{center}
\end{table}

\subsection{Meta-Initialization v.s. ImageNet-Initialization} 
\label{experiments:intra-dataset}




We next examine the full Algorithm \ref{meta-algo}, obtain higher-quality depth from the subsequent supervised stage, and \textbf{go beyond limited resources} to train on higher scene-variety datasets. 
Intuitively, higher scene variety may diminish meta-learning's advantages in few-shot/low-resource learning \cite{Chen_2021_ICCV}.
However, such studies are necessary to validate meta-learning for a realistic purpose since depth estimators are practically trained on diverse environments.
Comparison is drawn with baselines of the direct supervised learning without meta-initialization and begins from the standard ImageNet-initialization.


Table~\ref{table:intra-dataset} shows intra-dataset evaluation that trains/tests on each official data split.
Hypersim/NYUv2 evaluation is capped for depth at 20m/10m, respectively. 
Meta-Initialization attains \textbf{consistently} lower errors and higher accuracy than the baselines, especially AbsRel (-7.2\% on average). We display depth and 3D point cloud comparison in Fig.~\ref{intra-qualitative} and ~\ref{pointcloud}.
The gain simply comes from better training schema without additional data, constraints, advanced loss, or model design.


\subsection{Zero-Shot Cross-Dataset Evaluation} 
\label{experiments:cross-dataset}

\noindent\textbf{Protocol and evaluation}. To faithfully validate a trained model in the wild, we design protocols for zero-shot cross-dataset inference that also aims for sim-to-real purposes. High scene-variety and large-size synthetic datasets, Hypersim and HM3D, are used as the training sets. Real or more photorealistic VA, Replica, and NYUv2 serve as the test set, and their evaluations are capped at 10m. $D_{gt}$ and $D_{pred}$ are groundtruth and predicted depth. We use median-scaling in the protocol to compensate for different camera intrinsics, following common unsupervised or zero-shot depth evaluation protocols ~\cite{godard2017unsupervised,yin2021learning,zhang2023robust,Godard_2019_ICCV}, which computes a ratio $=\mathbf{median}$($D_{gt}$) / $\mathbf{median}$($D_{pred}$) and multiply it with the prediction to align with groundtruth scale in the evaluation. 
In Table \ref{table:cross-dataset-hm3d}, compared with the ImageNet-initialization, the meta-initialization \textbf{consistently} improves in nearly all the metrics, especially $\delta_1$ (on average +1.74 points). 


\noindent\textbf{Meta-Initialization as plugins}. To show wide applicability, we plug our meta-initialization into many recent high-performing architecture specialized for depth estimation, including BTS~\cite{lee2019big}, DPT (hybrid and large size) ~\cite{Ranftl2021}, DepthFormer~\cite{li2022depthformer}, AdaBins~\cite{bhat2021adabins}, GLPDepth~\cite{kim2022global}, and NDDepth~\cite{shao2023nddepth}. Comparison in Table \ref{table:cross-dataset-dedicated} shows using our meta-initialization consistently improves their original performances, giving them higher generalizability on zero-shot cross-dataset inference. Our meta-initialization is useful, especially since it only needs a few lines of code changes to modify learning schedules as a simple but effective piece.

\noindent\textbf{Comparison: single-image meta-learning depth}. Next, we compare with \cite{sun2022learn}, which also works on pure single images but requires multiple training datasets. They define tasks as different training set combinations and apply a consistency loss on the same scene but belonging to different tasks. We apply their strategy, train on NYUv2, HM3D, and Hypersim, define tasks as single or joint of two or three sets, compare with our meta-initialization trained on the three datasets, and test on VA. ConvNeXt-large is used. Results are shown in Table~\ref{table:learnAdapt}. We find the improvement by \cite{sun2022learn} is limited compared with HM3D$\to$VA in Table~\ref{table:cross-dataset-hm3d}. We think it is because scene structures in HM3D, NYUv2, and Hypersim differ greatly. Even for a dataset, different sequences contain varying spaces. This contrasts with their original domain on driving scenes, where most scenes are structurally similar, such as upper sky, lower road planes, and objects of cars or trees. Thus, their tasks by grouping training sets do not benefit much due to the lack of similarity in a task.

\noindent\textbf{Comparison: domain-adaptation depth}. We further compare to an unsupervised domain adaptation method T$^2$Net~\cite{zheng2018t2net} and semi-supervised domain adaptation methods ARC~\cite{zhao2020domain} and S2R-DepthNet~\cite{Chen2021S2RDepthNet}. These works adopt the setting SUNCG$\to$NYUv2.
We follow the training scripts by S2R-DepthNet and show the results in Table ~\ref{table:add1}.

\subsection{Better Depth Supervision in NeRF}
\label{experiments:nerf}
\begin{figure}[bt!]
    \centering
    \includegraphics[width=0.76\linewidth]{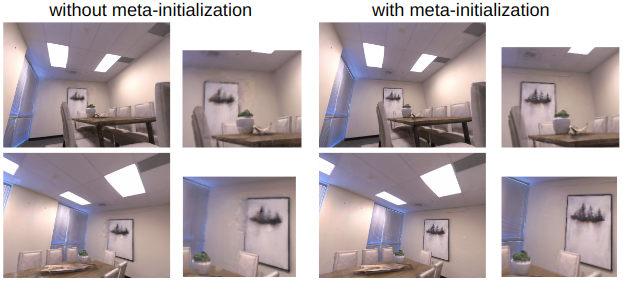}
    \vspace{-4pt}
    \caption{\textbf{Novel view comparison.} Zoom in for the best view.}
    \vspace{-21pt}
    \label{nerf_qual}
\end{figure}
We show that more accurate depth from meta-initialization can better supervise the distance $d$ a ray travels in NeRF. $d$ is determined by the volumetric rendering rule \cite{deng2022depth}. In addition to the vanilla pixel color loss, we use distance maps $d^*$, converted from monocular predicted depth by camera intrinsics, to supervise the training by $\mathcal{L}_D=|d^*-d|$. The experiment is conducted on Replica's office-0 environment with 180 training views. After 30K training steps, we obtain NeRF-rendered views and calculate the commonly-used image quality metrics (PSNR and SSIM, the higher the better). We use ConvNeXt-base to predict $d^*$. Table~\ref{nerf_metrics} shows a comparison made between with and without meta-initialization, and Fig.~\ref{nerf_qual} shows a visual comparison. 
\begin{table}[tb!]
\begin{center}
  \vspace{-3pt}
  \caption{\textbf{Results on depth-supervised NeRF.} We test on Replica 'room-0', 'room-1', room-2', 'office-0', 'office-1', and 'office-2' environments.}
  \vspace{-5pt}
  \ssmall
  \label{nerf_metrics}
  \begin{tabular}[c]
  {|
  p{1.2cm}<{\centering\arraybackslash}|
  p{0.8cm}<{\centering\arraybackslash}|
  p{0.8cm}<{\centering\arraybackslash}|
  p{0.8cm}<{\centering\arraybackslash}|
  p{0.8cm}<{\centering\arraybackslash}|}
  \hline
  & \multicolumn{2}{|c|}{\cellcolor[HTML]{99CCFF}w/o meta-initialization} & \multicolumn{2}{|c|}{\cellcolor[HTML]{F7BCDC}w/ meta-initialization} \\
       \multicolumn{1}{|c|}{\cellcolor[HTML]{F1F8B5} Environment}  &  \multicolumn{1}{|c|}{\cellcolor[HTML]{D5F5E3} PSNR$\uparrow$} &  \multicolumn{1}{|c|}{\cellcolor[HTML]{D5F5E3} SSIM$\uparrow$} & \multicolumn{1}{|c|}{\cellcolor[HTML]{D5F5E3} PSNR$\uparrow$} &  \multicolumn{1}{|c|}{\cellcolor[HTML]{D5F5E3} SSIM$\uparrow$}  \\
    \hline
       Room-0 & 29.99 & 0.818 & \textbf{30.92} & \textbf{0.837} \\
       Room-1 & 34.55 & 0.928 & \textbf{34.87} & \textbf{0.931}\\ 
       Room-2 & 36.68 & 0.956 & \textbf{37.46} & \textbf{0.961} \\ 
       Office-0 & 38.67 & 0.963 & \textbf{39.29} & \textbf{0.968} \\
       Office-1 & 36.20 & 0.943 & \textbf{36.87} & \textbf{0.946} \\
       Office-2 & 42.65 & 0.964 & \textbf{42.67} & \textbf{0.965}  \\
    \hline
  \end{tabular}
  \vspace{-20pt}
\end{center}
\end{table}

\vspace{-6pt}
\section{Conclusion and discussion}
\vspace{-3pt}
\label{conclusion}
This work investigates how meta-learning helps in pure single-image depth estimation and closely analyzes improvements.
Meta-learning can learn smooth depth from global context (\ref{experiments:scene-fitting}). It is a better initialization to obtain higher model generalizability verified on intra-dataset, zero-shot cross-dataset, and 3D representation evaluations (\ref{experiments:intra-dataset},  \ref{experiments:cross-dataset}, \ref{experiments:nerf}). From depth's perspective, this work studies an effective learning scheme to gain generalizability and validates with the proposed cross-dataset protocols. From meta-learning's perspective, this work proposes fine-grained task for a challenging pure single-image setting and studies a complex and practical goal of pixel-level real-valued regression.

\noindent\textbf{Discussion: large foundation models}.
Foundation models generally require a large corpus of pretrained data and still need a finetune set to adapt to the downstream task. In this case, another RGB-D dataset is required for adaptation.
In contrast, we only require an RGB-D set. If the size of a finetune set is not large enough, a large model may easily overfit by overparameterization and show worse generalization. We experiment finetuning foundation models using ConvNeXt-XXLarge and ViT-L/14 and compare with our meta-initialization. They are CLIP weights as initial parameters and further tuned on ImageNet22K. Finetuning those models does not win over our +Meta on Replica $\to$ VA and only show marginal gain on HM3D $\to$ VA.

\begin{table}[h]
\label{table:foundation}
\begin{center}
  \vspace{-13pt}
  \tiny
  \begin{tabular}[c]
  {
  p{1.2cm}<{\arraybackslash}
  p{2.3cm}<{\centering\arraybackslash}
  p{1.5cm}<{\centering\arraybackslash}
  p{2.1cm}<{\centering\arraybackslash}}
  \hlineB{2}
      AbsRel$\downarrow$ & ConvNeXt-XXLarge+finetuning & ViT-L/14+finetuning & ConvNeXt-Base+Meta \\
    \hline
      Replica  $\to$VA & 0.437 & 0.434 & \textbf{0.430} \\
      HM3D  $\to$VA & \textbf{0.160}  & 0.162 & 0.163 \\
    \hlineB{1}
    \hline
  \end{tabular}

  \vspace{-25pt}
\end{center}
\end{table}

{\small
\bibliographystyle{ieee_fullname}
\bibliography{egbib}
}

\end{document}